%% file: main.tex
\documentclass[conference]{IEEEtran}
\IEEEoverridecommandlockouts
\usepackage{cite}
\usepackage{amsmath,amssymb,amsfonts}
\usepackage{algorithmic}
\usepackage{graphicx}
\usepackage{textcomp}
\usepackage[table,xcdraw]{xcolor}
\usepackage{soul}
\usepackage{todonotes}
\usepackage{subcaption}
\usepackage{empheq}
\usepackage{hyperref}
\usepackage[T1]{fontenc}
\usepackage{tikz}
\usepackage{systeme,mathtools}
\usetikzlibrary{positioning,arrows.meta,quotes}
\usetikzlibrary{shapes,snakes}
\usetikzlibrary{bayesnet}
\tikzset{>=latex}

\begin{document}

\title{Generalising Discrete Action Spaces with Conditional Action Trees\\
% \thanks{ESRPC Grant thing}
}

\author{\IEEEauthorblockN{Christopher Bamford}
\IEEEauthorblockA{\textit{Game AI Group} \\
\textit{Queen Mary University of London}\\
London, UK \\
c.d.j.bamford@qmul.ac.uk}

\and
\IEEEauthorblockN{Alvaro Ovalle}
\IEEEauthorblockA{\textit{Game AI Group} \\
\textit{Queen Mary University of London}\\
London, UK \\
a.ovalle@qmul.ac.uk}
}
% \and
% \and
% \and
% \and
% \and

\maketitle

\begin{abstract}
There are relatively few conventions followed in reinforcement learning (RL) environments to structure the action spaces. As a consequence the application of RL algorithms to tasks with large action spaces with multiple components require additional effort to adjust to different formats. In this paper we introduce {\em Conditional Action Trees} with two main objectives: (1) as a method of structuring action spaces in RL to generalise across several action space specifications, and (2) to formalise a process to significantly reduce the action space by decomposing it into multiple sub-spaces, favoring a multi-staged decision making approach. We show several proof-of-concept experiments validating our scheme, ranging from environments with basic discrete action spaces to those with large combinatorial action spaces commonly found in RTS-style games.
\end{abstract}

\begin{IEEEkeywords}
action spaces, reinforcement learning, factorised policies, multi-agent, real-time strategy, actor-critic methods.
\end{IEEEkeywords}

\section{Introduction}

Training reinforcement learning agents to solve environments with large, complex action spaces is a notoriously difficult task~\cite{kanervisto_2020}. Several methods have been proposed to try to either reduce the space of actions by re-using model outputs for different action types \cite{masson_2016, vinyals_2017}, provide side information to facilitate the exploration of large numbers of possible actions \cite{samvelyan_2019, ontan_2013, justesen_2018}, or simplify the manipulation of the action spaces through action embeddings via mechanisms such as attention and graphs networks \cite{openai_2019, ma_2019, ammanabrolu_2020}. 
In this paper we propose a {\em Conditional Action Tree} as a paradigm to generalise several of these methods. {\em Conditional Action Trees} can be used to describe action spaces in a way that naturally reduces the required policy model output size whilst also allowing action parameterisation and action reduction using invalid action masking. We show how many of the action spaces frequently found in single, multi-agent and Real Time Strategy (RTS) games can be described using {\em Conditional Action Trees}. We also show that agents that have access to {\em Conditional Action Trees} as part of their state observations can learn high performing policies. We present several experiments where we purposefully modify the action space of a game environment to include several increasingly more complex features, whilst keeping the observation space and game mechanics consistent. In these experiments we show that agent operating with {\em Conditional Action Trees} maintains the performance of those operating with common action space constructions while significantly compressing the number of outputs, or \textit{logits}, required to furnish the policy distribution.

In addition to these experiments we also perform several ablation studies to show various possible modifications to the {\em Conditional Action Tree} formulation and how they can affect training. 

The results suggest that the {\em Conditional Action Trees} could offer an alternative to generically handle complex combinatorial action spaces with multiple components. As part of this work, {\em Conditional Action Trees} are made available for all environments in the Griddly Framework~\cite{bamford2020griddly}.

\section{Background}

In a discrete action setting, reinforcement learning (RL) has typically been adapted to environments with simple and small action spaces. Accordingly the implications that the size could have for the agent have been relatively overlooked. Let us start by considering a single actor-critic agent with a small repertoire of actions that consists in motion operations (e.g. up, down, left and right). In this setting, a policy will provide a probability distribution weighting each of the four directions. The agent then can sample this policy to select which action to apply to the environment. However these small manageable action spaces tend to be confined to either simple games or toy environments. The situation changes as we move to tasks requiring combinatorial actions such as in robotics~\cite{korenkevych_2019}, finance~\cite{yang_2020} and games that involve action spaces with several moving parts and interdependent components. For the latter, RTS games provide instances of actions spaces that can be particularly complicated. To consider a few examples, StarCraft II\cite{vinyals_2017}, $\mu$RTS~\cite{huang_2020} and BotBowl~\cite{justesen_2019} allow control of multiple individual units either by selecting their locations and then issuing commands to those units. Some of the units can perform certain types of actions that are not accessible to other units. Furthermore, some of those actions in turn require additional parameters. For instance, selecting a combat unit that can target several potential locations in the game requires to specify them. Moreover, the particular type of combat actions might be tied or dependent on the unit selected. Several techniques have been proposed to handle this kind of action spaces. \cite{kanervisto_2020} proposes several ways of shaping actions spaces and their relative advantages and disadvantages across several games. For the rest of this section we briefly review two strategies for action space shaping that have been recently proposed in the literature.

% NeuralMMO - \cite{suarez_2019}

\subsection{Parameterised Actions}

Parameterised action spaces commonly take the form of an action $a$ made from two components ${c_0, c_1}$ where the first component is a {\em type} of action and the second is a {\em parameter}. In \cite{masson_2016}, this action space shaping strategy was applied in the \textit{RoboCup 2D Half-Field-Offense} environment to beat the state-of-the-art hard-coded bots. The first action component defines whether the agent will \textit{dash, turn, tackle} or \textit{kick}. The second component defines continuous parameters for each of these actions. Four sets of parameters are used, however only one of them is used at each time-step depending on the action type selection. In larger environments such as RTS games, requiring parameters for every action quickly becomes infeasible as the number of action types increases. To contextualise the effect this can have for the size of the policy representation consider the example of \textit{BotBowl}. The game contains 17 action types that require an $x$ and $y$ position parameter. If we proceeded to parameterise the action space, 17 sets of $x$,$y$ positions would need to be predicted at each time step. From the point of view of an RL agent, the problem is exacerbated if we consider that the policy would have to specify each combination of $x$ and $y$ position. In a traditional \textit{BotBowl} map ($25\times5$) this would lead to a policy that requires to output $17 + 17\times25\times15 = 6,382$ logits (i.e. unnormalised scores) to parameterise these actions. The number grows exponentially with the map size, a $30\times30$ map for example, would require $17 + 17\times30\times30 = 15,317$ logits.

\subsection{Autoregressive Policies}

\begin{figure}
    \centering
    \input{images/autoregressive_pol}
    \caption{An autoregressive policy can be graphically represented as a directed acyclic graph where we can illustrate the dependency of a component $c_k$ on the previous components $c_{<k}$.}
    \label{fig:autoregressive-policy}
\end{figure}
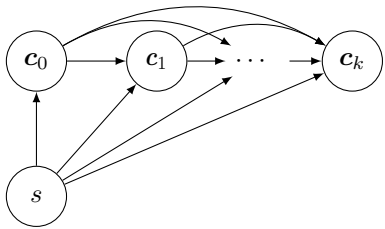

The curse of dimensionality and the combinatorial explosion faced in certain action spaces renders the typical action selection approach highly impractical. An alternative comes from reflecting on the structural relations that exist in complex action spaces. For example, when comparing among all the potential actions that an agent could choose to enact, not all of them will belong to the same level of abstraction. Some actions will be more self-contained, whereas others may need a group of actions to be properly contextualised. Actions can also manifest to an agent as affordances, that is, arising from its coupling with the environment at a particular moment. Moreover, some actions also exert some degree of influence on each other, for instance mutual exclusivity or forming other types of associations. 

It is possible to capture some of these notions more concretely by representing a policy in a more expressive manner. In \cite{vinyals_2017} the authors suggest an autoregressive model of the form:

\begin{equation}
    \pi(a|s) = \pi(c_0,\dots,c_k|s) =\prod_{k=0}^K{\pi(c_k|c_{<k}, s)}
\end{equation}

to decompose the action space into a sequence of sub-spaces. Instead of obtaining $a$ in the full action space, an agent samples multiple sub-actions or components $c_k$ that depend on the previous $c_{<k}$ choices (illustrated in Fig. \ref{fig:autoregressive-policy}). \cite{vinyals_2017} explores the usage of conditional policies within the context of StarCraft II. However they relax the constraints imposed by the autoregressive model opting for a policy $\pi(a|s) = \prod^K_k \pi(c_k|s)$. In \cite{vinyals_2019}, the approach is extended substantially as the architecture considers a conditional policy that captures the context of previous actions through different embeddings. The action sub-space decomposition is facilitated by an \textit{invalid action masking} scheme to prevent the agent from selecting actions that are invalid or cannot be performed in the current state, where a function identifier $c_0$ determines the number of subsequent function arguments $c_1 \dots c_k$. From the point of view of the implementation and the capacity required by a policy, this form of decomposition implies a significant reduction in the number of actions that are effectively considered. Here we also take an approach that places an invalid action masking scheme as a crucial component to tackle the issue of policy decomposition and describe it in more detail in Section \ref{sec:invalid-action-masking}.

\section{Conditional Action Trees} \label{sec:conditional-action-trees}

Conditional action trees (CAT) offer a generalisation of discrete action spaces to provide an interpretation of action selection as the process of traversing along a chained sequence of action components with different levels of dependency. To complete the characterization of a {\em Conditional Action Tree} we first need to define three main elements: {\em Action Trees}, {\em Valid Action Trees}, and finally {\em Conditional Masking}.

\subsection{Action Trees} \label{sec:action-trees}

\begin{figure}
    \centering
    \includegraphics[width=\linewidth]{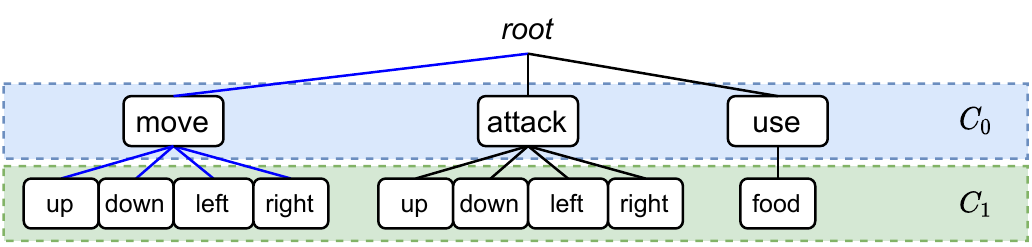}
    \caption{An action tree consisting of nine possible actions and two components $C_0 = 3$ and $C_1 = 4$. The possible action are move or attack in any of the four directions, attack in any of the four directions and finally to use a food item. {\em Move}, {\em Attack} and {\em Use} cannot be performed at the same time.}
    \label{fig:action-tree}
\end{figure}

We start by formulating a single action as a list of a fixed number of components $a = \{c_0, c_1 ... c_n\}$, where each component is a discrete value taken from a set of possible values for that particular component $c_k \in C_k$. Actions in the same component level are mutually exclusive. For example, \textit{move left} and \textit{move right} must be options within a single component $C_k$. The possible values of $C_k$ are determined first, by the specification of environment, and second by the values of previous selections $C_k = f(c_0, c_1, ... , c_{k-1})$.

These restrictions naturally allow the components to form a tree structure, where a path from the root node to any leaf forms the action. An example of an {\em action tree} is shown in Fig.~\ref{fig:action-tree}. Note that under this specification an environment requiring the agent to specify a single atomic action at each time step results in an {\em action tree} with a single component, $a = \{c_0\}$. Parameterised action spaces that contain an action type and a discrete action parameter can also be described by action trees with two components, $a = \{c_0, c_1\}$.

Previous work has touched upon the idea of using trees as a formalization of action spaces with multiple components such as in \cite{fan_2019}, where the tree structure is referred to as a {\em Hierarchical Action Space}. Other works have used action spaces that are similar to those used in this paper as examples of  action trees. The {\em Global Action Space} in \cite{huang_2019} for example can also be described as an {\em action tree}. 

\subsection{Valid Action Trees} \label{sec:valid-action-trees}

We define a \textit{valid action tree} as a sub-tree of an action tree at a particular environment state, where the nodes of the sub-tree correspond to {\em possible} actions in that state. For example, consider the tree in Fig.~\ref{fig:action-tree}, an agent in a state where there are no enemies surrounding it and does not have food in its inventory has a {\em valid action tree} only consisting of the left-most {\em move} branch and its children.

In the context of reinforcement learning, a valid action tree is provided by the environment at each time step. Valid action trees are then used to construct the {\em Invalid Action Masks} which are described next in Section~\ref{sec:invalid-action-masking}. These masks index the child nodes that are available in the full action tree.

\subsection{Invalid Action Masking} \label{sec:invalid-action-masking}

Invalid action masking (IAM) \cite{huang_2020} is a technique used to stop agents from sampling actions that are invalid in a particular game state. IAM is useful in environments where the action space is large, and some of the actions are only available in certain states. For example in RTS games \cite{vinyals_2019, vinyals_2017, samvelyan_2019}, the agent's action may consist of selecting a unit or units from a large list, and then issuing commands to those units. The commands sent to those units can also be unique to particular unit types. This results in a large number of options in the action space that are invalid. In policy gradient and actor critic methods in deep RL, IAM is applied to the logits, $\textbf{l} \in \mathbb{R}^n$, produced by a neural network by replacing the logits corresponding to invalid actions with large negative numbers. This forces the probability of selecting those actions to tend towards 0. 

For instance, let us assume a compound policy constituted by $K$ independent components, such that $\pi(a|s) = \prod^K_k \pi(c_k|s)$. This type of action policy could be described by:

\begin{equation}
    \pi(a|s) = [\pi(c_{0}|s), \pi(c_{1}|s), \dots, \pi(c_{k}|s)]
    \label{eq:indep_pol}
\end{equation}

For each of the components in equation \ref{eq:indep_pol}, a value is selected following a softmax sub-policy. We can create a mask to modify the logits to assign large negative numbers to actions deemed as non-viable or inaccessible. The modified logits result in $\hat{\textbf{l}} = \textbf{l} + \textbf{m}$ where $-\infty < m_i \ll 0$. It then follows that the {\em masked} logits alter the probability of a value of $c_i$ of being sampled:

\begin{equation*}
    \pi(c_i|s) = 
    \begin{cases}
        0 & \text{if } m_i \longrightarrow - \infty \\
        \frac{e^{l_i}}{\sum_j^N{e^{l_j}}}  & \text{if } m_i \quad=\quad0
    \end{cases}
\end{equation*}

%the unmasked softmax sub-policy of any given $c_i$ corresponds to,

% 
% For clarity we now drop the time subscript $t$ from a component $c_{i,t}$. From equation \ref{eq:indep_pol}, the unmasked softmax sub-policy of any given $c_i$ corresponds to,

% \begin{equation}
%   \pi(c_i|s) = \frac{e^{l_i}}{\sum_j^N{e^{l_j}}}
%   \label{eq:softmax}
% \end{equation}

% We can create a mask to modify the logits to assign large negative numbers to actions deemed as non-viable or inaccessible:

% \[
%     \textbf{m} \in \{-\infty,0\}^n \quad\quad
%     \hat{\textbf{l}} = \textbf{l} + \textbf{m}
% \]

% The masked logits $\hat{\textbf{l}}$ are then used in the calculation of the policy modifying equation \ref{eq:softmax}:

% \begin{equation}
%     \pi(c_i|s) = \frac{e^{\hat{l}_i}}{\sum_j^N{e^{\hat{l}_j}}}
% \end{equation}

 In PySC2 \cite{vinyals_2017}, $\mu$RTS \cite{ontan_2013} and BotBowl\cite{justesen_2019} action masks can be constructed from lists of available actions that are provided by the environment implementations, however, these action masks do not take into account that the masking of some sub-actions can depend on the sampled values of others. As an example, in an environment with units that are selected by coordinates and the set of available actions for each unit is disjoint, the mask for the available actions is dependent on the selection of the unit. Masks that are naively constructed using these lists can still lead to select actions that are not available, as the list does not take into account the selection of the unit. \cite{huang_2021} introduces a two-step method for generating masks where the unit location is selected using masked logits and then a second mask is generated based on that selection. This significantly improves training as the mask for unit actions is dependent on the selected unit.

\subsection{Conditional Masking}
\label{sec:cond_mask}

The two-step method of masking in \cite{huang_2021} can be generalised to an n-step masking method when the environment provides a valid action tree as described in Section~\ref{sec:valid-action-trees}. We refer to this generalization of action selection and masking as a \textit{Conditional Action Tree (CAT)}.

A CAT is constructed by adding a mask at each node of a valid action tree, defining which child nodes of the complete action tree are available. 
An action is constructed by starting at the root node of the valid action tree and selecting a child node from the masked distribution. This child node contains the mask to use for the next component. Thus, first the mask is obtained as $\textbf{m}_{k+1} \sim p(\textbf{m}_{k+1}|c_k)$, to produce a masked sub-policy to sample a component $c_{k+1} \sim p(c_{k+1}|\mathbf{m}_{k+1},s)$. This process continues until all action components have been sampled. The full compound policy, as illustrated in Fig. \ref{fig:masked_pol}, is factorised as:

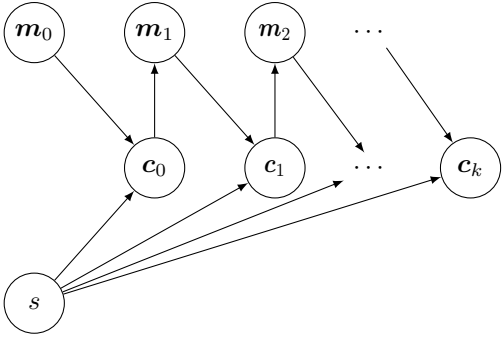
\begin{figure}
    \centering
    \input{images/mask_dep_pol}
    \caption{A graphical model representing the policy as a joint distribution of masks $m$ and components $c$. In a CAT, a component $c_0$ is sampled from the options allowed by the mask $\textbf{m}_0$. The next mask $\mathbf{m}_1$ depends on $c_0$ which in turn constrains the next possible component $c_1$. The process is repeated until all $c_k$ have been sampled.}
    \label{fig:masked_pol}
\end{figure}

\begin{equation}
    \pi(a|s) = p(\mathbf{m}_0) \prod_k^K p(\mathbf{m}_{k+1}|c_k)p(c_k|\mathbf{m}_k,s)
\end{equation}

\section{Actor-Critic with Conditional Action Trees} \label{sec:policy-gradient-methods}

\subsection{IMPALA}

The description of the action spaces provided by CAT is naturally agnostic to the choice of the RL algorithm. We examine this perspective within the context of IMPALA, an actor-critic based framework introduced in \cite{espeholt_2018}. Unlike A3C \cite{mnih2016} or other similar distributed approaches where the agents share their gradients, IMPALA considers the acting and the data collection as independent from the learning step. That is, it separates the learners who are in charge of computing the gradients and sharing the most recent parameters, from the actors whose role is to execute a policy, only sharing back with the learners the observations gathered during an episode. 

\subsection{V-trace and masking}

As an actor-critic, IMPALA learns $V_\theta(s)$ parameterised by $\theta$ to be used as part of the baseline, and a policy $\pi_\phi$ parameterised by $\phi$. Each actor executes their own policy $\mu$ by retrieving the latest policy $\pi$ from the learner. Meanwhile the learner updates continuously the parameters $\theta$ and $\phi$. As the process occurs in parallel and in a decoupled manner, there will be a discrepancy between the policy $\mu$ from an actor and $\pi$. Namely, the trajectories $(s_t,a_t,r_t \dots)$ collected by an actor come from a policy $\mu$ that has become obsolete with respect to $\pi$. IMPALA proposes to address these off-policy corrections by introducing a v-trace target,

\begin{equation}
    v_t  = V(s_t) + \sum_{i=t}^{t+n-1} \gamma^{i-t} \big(\prod_{j=t}^{i-1} u_j\big) \delta_i V
\label{eq:v-trace}
\end{equation}

where $\delta_i V$ corresponds to a temporal difference term,

\begin{equation*}
    \delta_iV = \rho_i (r_i + \gamma V(s_{i+1}) - V(s_i))
\end{equation*}

the v-trace adjusts the weight of the contributions provided by the actors through the presence of two truncated importance sampling weights $\rho_i=\min(\overline{\rho},\frac{\pi}{\mu})$ and $u_j=\min(\overline{u},\frac{\pi}{\mu})$. Thus the second part of v-trace target acts as a correction term. For example assuming $\overline{p}$ and $\overline{u} \geq 1$, if $\mu>\pi$ the learner would downweight the observations and actions followed by the actor. Intuitively, if this ratio tends towards a low number it indicates that the policies have diverged significantly. The extent to which more recent $\delta_i V$ affect the update of a previous $V$ is captured by the product of $u_{t:i-1}$ where $\overline{u}$ serves as a hyperparameter controlling the convergence speed towards $V$. In turn, $\rho$ determines to which $V$ we converge. A $\overline{\rho}$ close to 0 leads convergence towards a $V^\mu$ as the correction term becomes negligible in the v-trace target.

It is important to note that for CAT we do not just consider a single set of importance sampling weights $\{\rho, u\}$ but instead we must account for multiple corrections dependent on the various sub-policies such that $\rho_{k,i} = min(\overline{\rho}, \frac{\pi(c_k|m_k,s)}{\mu(c_k|m_k,s)})$ and $u_{k,j} = min(\overline{u}, \frac{\pi(c_k|m_k,s)}{\mu(c_k|m_k,s)})$ for a sub-policy $k$. Moreover, we must synchronize the masks applied to $c_k$ in both $\pi$ and $\mu$. Similarly, for updating the policy parameters $\phi$ we adapt,

\begin{equation*}
    \rho_{k,i}\nabla_\phi \log \pi_\phi(c_k|m_k, s)(r_t + \gamma v_{t+1} - V_\theta (x_t))
\end{equation*}

to consider the inclusion of the masks and to propagate the gradients to all sub-policies.

\section{Experiment Setting} \label{sec:experiment-setting}

\begin{figure*}
    \centering
    \includegraphics{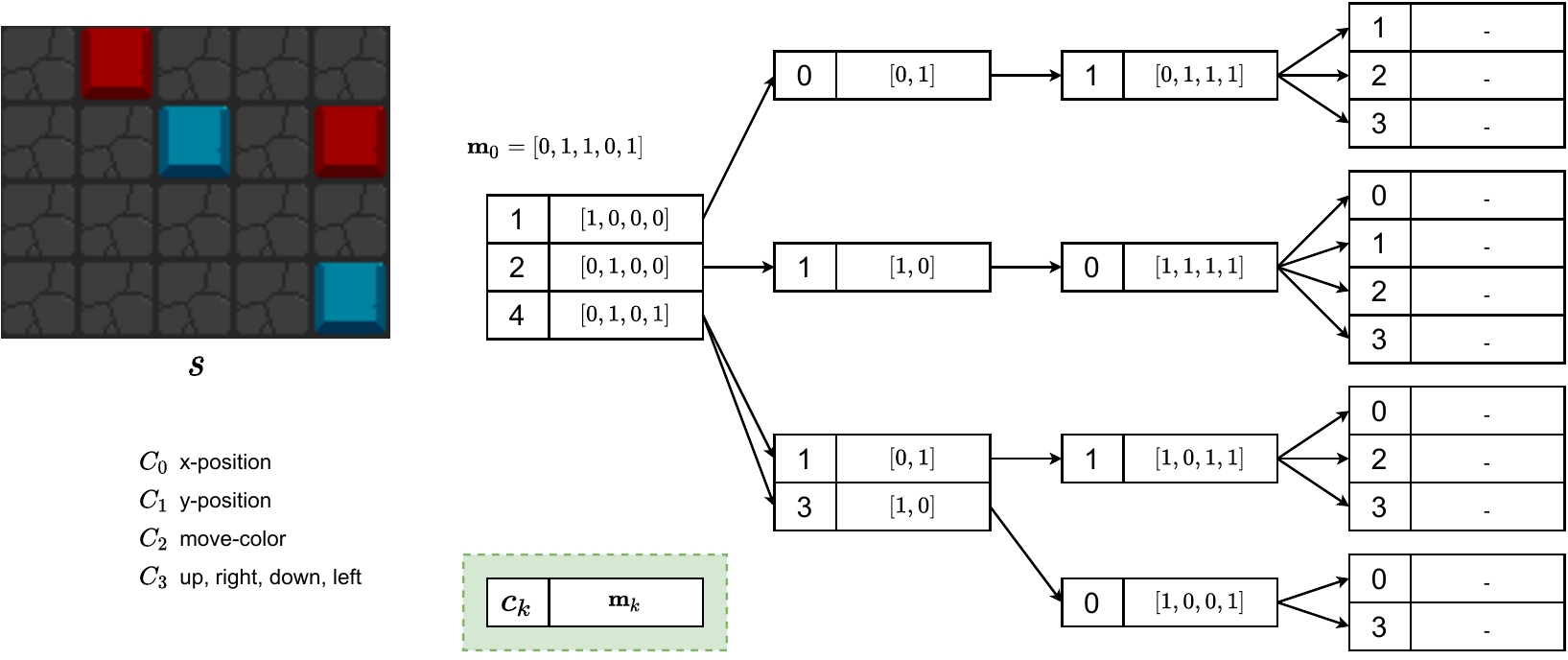}
    \caption{Image of a conditional action tree from a 5x4 {\em Clusters} level configured in with the MSa action space as described in Section~\ref{sec:action_space_variations}. The agent is configured with an action space with 4 components, the agent selects which object to move by it's position and it's colour, it can then choose which direction to move the {\em box}. The CAT shown contains the selected action component $c_k$ and the mask $\textbf{m}_k$ for each possible valid combination of components.}
    \label{fig:conditional-action-tree}
\end{figure*}

\subsection{The "Clusters" Game} 

\begin{figure}
    \centering
    \begin{subfigure}{.5\linewidth}
        \centering
        \includegraphics[width=.9\textwidth]{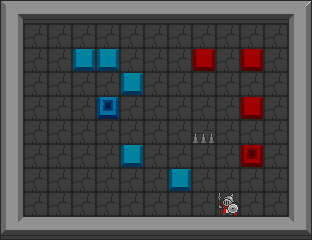}
        \caption{Global}
    \end{subfigure}%
    \begin{subfigure}{.2\linewidth}
        \centering
        \includegraphics[width=.9\textwidth]{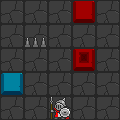}  
        \caption{Agent}
    \end{subfigure}%
    \caption{An example of a level in the Clusters game, showing (a) the entire game and (b) the viewpoint of the agent.}
    \label{fig:clusters}
\end{figure}

We perform our experiments in the {\em Clusters} environment provided by Griddly~\cite{bamford2020griddly}. {\em Clusters}\footnote{\url{https://griddly.readthedocs.io/en/latest/games/Clusters/index.html}} is a game in which coloured {\em boxes} must be clustered together in specific locations defined by the environment level. The environment contains five levels with a set of movable coloured {\em boxes} and a single fixed-position {\em block} of each colour. The agent receives a reward of +1 each time it pushes a coloured {\em box} against a fixed location {\em block} of the same colour. When a coloured {\em box} is pushed against its respective {\em block}, it becomes a {\em block} itself. If all {\em boxes} are converted to {\em blocks} the episode is completed successfully. Some levels also contain {\em spikes} which give the player a negative reward (-1) and terminate the episode if the agent or any {\em boxes} collide with them.

The observation space of the agent consists of a $5\times5$ grid where the agent itself is situated at the center-bottom of the grid as shown in Fig.~\ref{fig:clusters}. Each cell of the $5\times5$ grid contains 10 binary values describing whether an object is present in each cell. The 10 objects are as follows: three (red, green, blue) coloured {\em boxes} and three associated {\em blocks}, {\em walls}, {\em spikes}, the agent and finally a {\em broken box} which only appears in the final state of an episode if a coloured {\em box} is pushed against {\em spikes}.

\subsection{Action Space Variations} \label{sec:action_space_variations}

By default, the agent's movement is restricted to moving forward one position, or rotating $\pm 90$ degrees every step. {\em Boxes} are "pushed" by the agent when the agent attempts to move into the cell occupied by the {\em box}.

In our experiments, we modify these action spaces to make it increasingly more complex whilst keeping the game mechanics, observation space and reward scheme consistent. This allows us to test the Conditional Action Tree formulation on different action spaces with minimal influencing factors. The only significant change we make to the environment across experiments is when we remove the avatar and allow the agent to move {\em boxes} independently by selecting their $x$ and $y$ coordinates. These action space variations are explained below:

\subsubsection{\textbf{Move (M)}}

The first action tree variation is the default action space provided by the {\em Clusters} environment. The action space consists of rotate left, right and move forward. As mentioned in Section~\ref{sec:action-trees}, this is equivalent to an {\em Action Tree} with a single component $a = \{c_0\}$, with $c_0 \in {0,1,2,3}$.

\subsubsection{\textbf{Move + Push (MP)}}

Next we modify the action space to consider that the agent can no longer {\em push} boxes by simply moving into the location occupied by them. We define a separate {\em push} action that must be performed in order to move any of the boxes. The {\em push} action has no effect unless there is a box directly in front of the agent. The {\em move} action is left unmodified, other than the fact that it can no longer be used to push boxes. As the {\em move} and {\em push} actions are mutually exclusive they are confined to the first level in the tree $C_0 = \{0,1\}$, whilst the second component $C_1$ contains either the three move parameters or the single {\em push} parameter.

\subsubsection{\textbf{Move + Push + Separate colours (MPS)}}

This action space configuration contains the same modifications as the MP variant, however it splits the {\em push} component into three to account for the separate colours. The agent must select the correct {\em push} action, depending on which colour {\em box} it is pushing (i.e. push green, push blue, push red). Similarly to {\em MP}, the action space consists of two components, but the first one now contains the three different push actions instead of one, that is $C_0 = \{0,1,2,3\}$. The second component $C_1$ remains the same.

\subsubsection{\textbf{Move - Agent (Ma)}}

To make the action space significantly larger we remove the agent and the associated ego-centric partial observability. Thus the input consists of the entire $13\times10$ grid with the same 10 binary digits per cell. The {\em boxes} are now moved first by selecting their $x$ and $y$ coordinates and then by issuing the direction where to move it. This action space has three components: $C_0 = \{ \text{valid x coordinates}\}$, $C_1 = \{ \text{valid y coordinates} \}$ and $C_2 = \{0,1,2,3\}$ referring to the movement directions {\em up}, {\em down} {\em left} and {\em right}.

\subsubsection{\textbf{Move + Seperate colours - Agent (MSa)}}

The final and largest action space we consider starts with the same formulation as {\em Ma}, but separates the colour components in the same way as done in {\em MPS}. This results in an action space with four components: x, y, action type and action parameters. An example of a conditional action tree for this space is shown in Fig.~\ref{fig:conditional-action-tree}

\subsection{Baselines} \label{sec:baselines}

For each of the variations of the action space described in the previous section, we compare against two baselines which are designed to show the benefits and limitations of the CAT paradigm. The baselines modify only the way that the model interacts with the action space in terms of number of logits required. The number of actions and mechanics of the game are consistent.

\subsubsection{\textbf{No Masking}}

For the first comparison we use the same action components as a CAT but remove the Invalid Action Masking entirely. This means that the component selections are made independently of each other and invalid actions can be selected.

\subsubsection{\textbf{Depth-2}} \label{sec:depth-2} 

The second comparison also uses a conditional action tree structure, but flattens the action tree to only a depth of two. The separate $x$ and $y$ components (only available in MSa and Ma) are flattened into a single $xy$ component. Additionally the action type and action parameter components are flattened into a single selection. This flattening process was also considered in \cite{kanervisto_2020} where multi-discrete actions are flattened into single discrete spaces. Table~\ref{tab:policy-logits} shows the number of logits per-component for all experiments and the equivalent number of logits required in the depth-2 representation.

\begin{table}[]
    \centering
    \begin{tabular}{lrlllr}
        \multicolumn{1}{l|}{}    & \multicolumn{1}{c}{$|C_0|$} & \multicolumn{1}{c}{$|C_1|$}  & \multicolumn{1}{c}{$|C_2|$}  & \multicolumn{1}{c|}{$|C_3|$}                 & \multicolumn{1}{r}{Total Logits} \\ \hline
        \multicolumn{1}{l|}{\textbf{M}}   & 3                      & \cellcolor{lightgray}         & \cellcolor{lightgray}      & \multicolumn{1}{l|}{\cellcolor{lightgray}} & 3                                \\
        \multicolumn{1}{l|}{\textbf{MP}}  & 2                      & \multicolumn{1}{r}{3}         & \cellcolor{lightgray}      & \multicolumn{1}{l|}{\cellcolor{lightgray}} & 5                                \\
        \multicolumn{1}{l|}{\textbf{MPS}} & 4                      & \multicolumn{1}{r}{3}         & \cellcolor{lightgray}      & \multicolumn{1}{l|}{\cellcolor{lightgray}} & 7                                \\
        \multicolumn{1}{l|}{\textbf{Ma}}  & 13                     & \multicolumn{1}{r}{10}        & \multicolumn{1}{r}{4}      & \multicolumn{1}{l|}{\cellcolor{lightgray}} & 27                               \\
        \multicolumn{1}{l|}{\textbf{MSa}} & 13                     & \multicolumn{1}{r}{10}        & \multicolumn{1}{r}{4}      & \multicolumn{1}{r|}{3}                     & 30                               \\
        \multicolumn{6}{c}{\textbf{Depth-2}}                                                                                                                                                              \\
        \multicolumn{1}{l|}{\textbf{M}}   & 3                      & \cellcolor{lightgray}         & \cellcolor{lightgray}      & \multicolumn{1}{l|}{\cellcolor{lightgray}} & 3                                \\
        \multicolumn{1}{l|}{\textbf{MP}}  & 4                      & \cellcolor{lightgray}         & \cellcolor{lightgray}      & \multicolumn{1}{l|}{\cellcolor{lightgray}} & 4                                \\
        \multicolumn{1}{l|}{\textbf{MPS}} & 6                      & \cellcolor{lightgray}         & \cellcolor{lightgray}      & \multicolumn{1}{l|}{\cellcolor{lightgray}} & 6                                \\
        \multicolumn{1}{l|}{\textbf{Ma}}  & 130                    & \multicolumn{1}{r}{4}         & \cellcolor{lightgray}      & \multicolumn{1}{l|}{\cellcolor{lightgray}} & 134                              \\
        \multicolumn{1}{l|}{\textbf{MSa}} & 130                    & \multicolumn{1}{r}{12}        & \cellcolor{lightgray}      & \multicolumn{1}{l|}{\cellcolor{lightgray}} & 142
    \end{tabular}
    \caption{This table shows the number of action components, their sizes in term of number of logits and the total logits needed in the policy output for the action space variations described in Section~\ref{sec:action_space_variations}. We also show the number of logits that are required in the {\em Depth-2} model.}
    \label{tab:policy-logits}
\end{table}

\subsection{Masking Ablation}

To show that structure of the tree and the resulting conditional masking has an effect on the learning of the policy, we perform an experiment where we relax the conditional masking restrictions and compare it against the fully conditional masking.
We relax the conditional masking of the tree by collapsing the masking across the tree breadth-wise, so all masking is effectively a union of all the possible masks at each depth. This method is equivalent to applying a single mask to the entire action space with no consideration for dependencies between action selections. We refer to the relaxed {\em Collapsed} and full {\em Conditional} masking options in further sections as CAT\_CL and CAT\_CD respectively.

\subsection{Model Architecture}

We keep the model architecture consistent throughout all experiments as much as possible. The size of the model input observations differs between partially observable agent-based environments (\textbf{M}, \textbf{MP}, \textbf{MPS}) and unit-selecting environments (\textbf{Ma}, \textbf{MSa}). The partially observable environments have a $5\times5\times10$ observation space, while for the unit-selecting environments it is $13\times10\times10$. Additionally, the final layer in each experiment outputs the number of logits shown in Table~\ref{tab:policy-logits}. The model itself contains two convolutional layers with padding 1 and kernel size 3 that up-scales the number of values in each channel to 32 and then 64 respectively, whilst keeping the width and height the same. After these layers, the output tensor is flattened and then passed through two linear layers with 1024 and 512 neurons. We then use a separate actor and critic head. The actor head contains a further two linear layers, first to compress to 256 nodes and then a final layer to output predicted logits. The critic head contains a single layer which outputs the single predicted value.

\section{Results}

\begin{figure*}
    \centering
    \includegraphics[width=.9\linewidth]{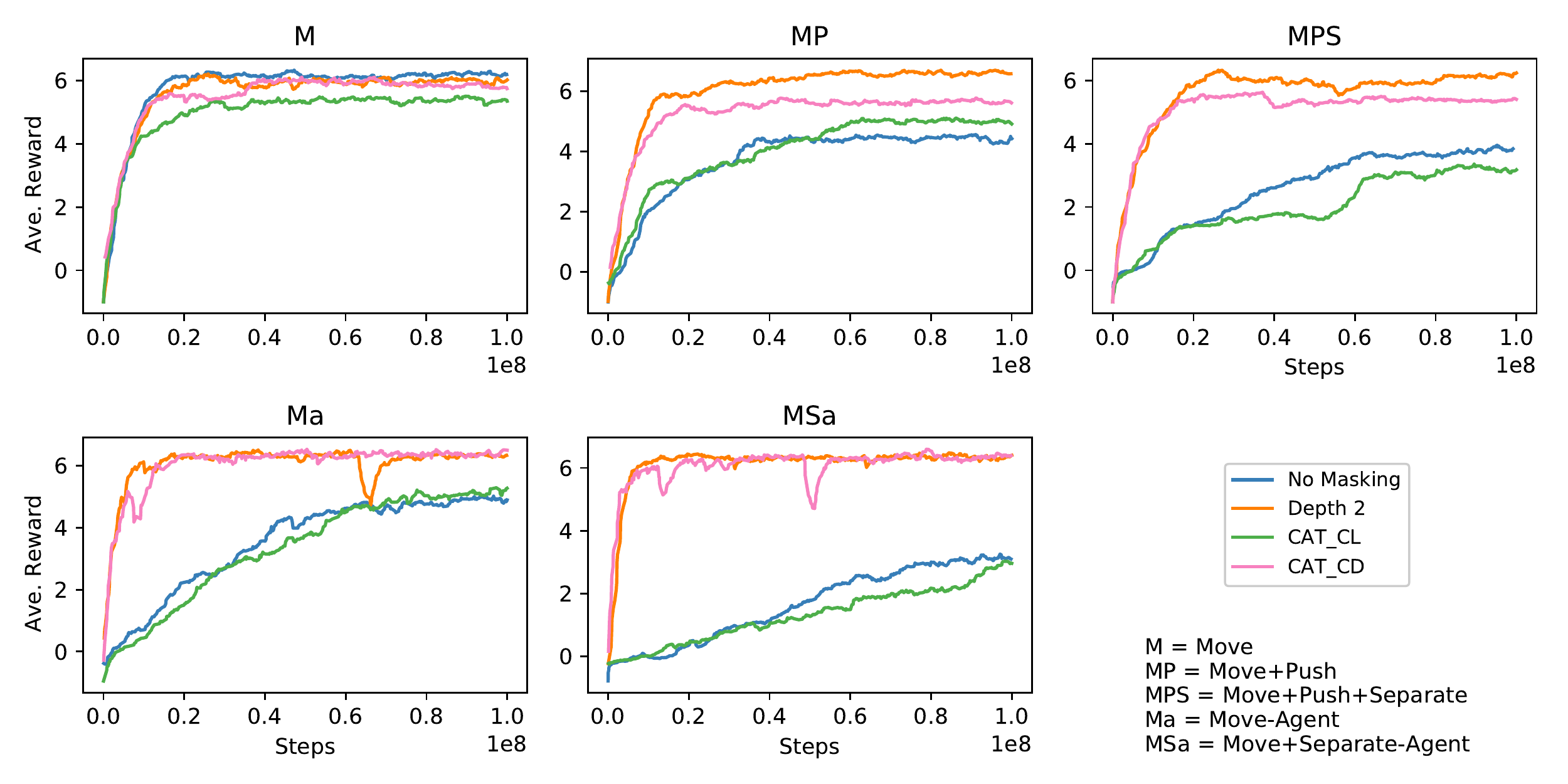}
    \caption{The average episode reward during training of the 5 different action space variations as described in Section~\ref{sec:action_space_variations}. For each of the 5 action space variations, we compare three policies with the same action tree structure, but different masking methods: No Masking, CAT\_CD (conditional) and CAT\_CL (collapsed). We also provide a comparison with a model policy that uses an action tree limited to depth 2 as described in Section~\ref{sec:depth-2}}
    \label{fig:results}
\end{figure*}

In total we run 4 experiments on each variation of the action space of the \textit{Clusters} game. The four experiments contain the two baselines as previously described, and two versions of masking (CAT\_CL and CAT\_CD).

The first variation \textbf{M} provides evidence that the formulation of conditional action trees generalise to simple action spaces. In this environment, all variations of the action space are almost identical and therefore have similar performance. Masks in this environment have little effect because only a few actions are ever invalid. \textbf{MP} and \textbf{MPS} variations begin to show that the fully conditional tree CAT\_CD and the depth-2 action tree policies learn faster and plateau at high-scoring policies. Depth-2 action policies in these variations are in fact slightly better performing than the more hierarchical formulation of the Conditional Action Tree, in addition of using one less logit in their policies. The reason for this is that in the \textbf{MP} and \textbf{MPS} the structure of the associated action tree has a degree of 1 in all of the {\em push} nodes, making the tree structure redundant for the push actions. In cases like these, where parent nodes have only single children, it is more efficient to flatten these nodes into a single set of children.

Conditional Action Trees excel in the variations with the highest branching factors. \textbf{Ma} and \textbf{MSa} both require the policy to select an individual unit to perform an action at each time step. As expected, the depth-2 policy and CAT\_CD have similar performance as they are both CATs, but CAT\_CD splits the $x$, $y$ selection into separate components, which results in a greater than 4x reduction in the number of logits required by the policy, with no loss in performance.  

The results for experiments on all five test environments with {\em Collapsed} (CAT\_CL) masks are also shown in Fig.~\ref{fig:results}. We can see that with {\em Conditional Action Trees} the full {\em Conditional} (CAT\_CD) masking strategy is important for efficient training, as the {\em Collapsed} masking strategy performs similarly to the {\em No Masking} Baseline. 

\section{Discussion}

Trees are a useful data structure across many fields of computer science, and can provide a natural representation for action spaces. Although the formulation and the experimental setting focused on discrete action spaces, we hypothesize that in principle the formulation can be extended to continuous spaces by implementing a similar parameterisation to structure the action components associated to specific densities. To the best of our knowledge this direction has not been explored.

It is important to note that the degree of subtrees in a CAT should be taken into consideration when deciding on parts of the tree that could be flattened, as this can lead to unnecessary increase in policy size. 

The current work presented the CAT formulation in five toy scenarios intended to recreate, with different levels of complexity, the conditions frequently exhibited in various single, multi-agent and RTS games. Further work will be required to analyse the behavior of the CAT in more complex domains such as $\mu$RTS or BotBowl. Part of the current limitation resides in adapting these and other environments to provide \textit{Valid Action Trees}. With a {\em Conditional Action Tree} the parameterised part of BotBowl's action space could be reduced from $6392$ logits to $25+15+17 = 57$

Other relevant research on how to handle large action spaces has applied techniques such as evolutionary algorithms \cite{baier_2018, justesen_2018}. These proposals have also been tested in scenarios that require multiple actions per time-step. A naive approach to work with CATs within this context would be to recursively append the tree to its own leaf nodes, resampling until a condition specifying the required number of actions is fulfilled.

In its current form a CAT makes specific assumptions about the conditional dependencies between actions (Section \ref{sec:cond_mask}). Following \cite{vinyals_2019}, a potential future research avenue is to explore the possibility of modelling more complex dependencies. Namely, by contextualizing further the selection of a $c_k$ with with an encoding learned from previous components $c_{<k}$.

\subsection{Entropy reduction}

The process by which an agent acquires a policy in reinforcement learning is essentially an exercise in reducing its behavioral uncertainty. Although it remains important that the agent maintains a level of flexibility that can support an adequate generalization \cite{ziebart2008,haarnoja2017}. It must effectively has to be able to discriminate the actions that are beneficial from those that are not in a particular state. We can interpret this procedure of action discrimination as a process of entropy reduction, as an initial high entropy policy is transformed by redistributing the probability mass or density to weight those actions that have been identified as more favorable. For large action spaces it is evident that this process becomes more complex as the behavioral possibilities explode. The view posited by CAT is that it is possible to exploit the structure of the action space to facilitate the acquisition of behaviorally relevant policies as we can state that for two arbitrary segments of the action space, $C_i$ and $C_j$, the mutual information between them is $I(C_i;C_j) \geq 0$. This implies that there can be information to be gained by using existing relationships between actions. To express it differently, conditioning guarantees us that $H(C_i|C_j) \leq H(C_i)$ with equality iff $p(c_i,c_j) = p(c_i)p(c_j)$. Thus constructing an action tree becomes a tool that contributes to the process of entropy reduction at the level of the policy as it decomposes what is potentially a large flat action space into multiple smaller sub-spaces. 

\section{Conclusion}

In this paper we have proposed a formalisation of a tree structure for representing discrete action spaces with any number of components. We have provided the required steps to adapt already existing action spaces to conform to a {\em Conditional Action Tree}. From a technical perspective, a side effect of imposing a structure to the action space is the reduction of the elements considered by a policy. The experiments showed that this modification does not reduce the sample efficiency during training and achieves comparable performance while resulting in significantly smaller models with less parameters.

As part of the this work, Griddly~\cite{bamford2020griddly} implements built-in functionality for generating {\em Valid Action Trees} and we provide all reproducible examples in a github repository~\footnote{\url{https://github.com/Bam4d/conditional-action-trees}}. We also provide all training parameters, statistic and videos using Weights and Biases~\footnote{\url{https://wandb.ai/chrisbam4d/conditional_action_trees}}. 

We encourage the developers of reinforcement learning environments, especially those with large discrete action spaces to provide {\em Valid Action Tree} observations in their environments.

\section*{Acknowledgment}

We would like to thank Paulo Rauber, Simon Lucas and Shengyi Huang for their valuable feedback and to the ITS Research team at Queen Mary University of London for providing technical support necessary to run the simulations. This research utilised Queen Mary's Apocrita HPC facility, supported by QMUL Research-IT. \url{http://doi.org/10.5281/zenodo.438045}. 

\bibliographystyle{IEEEtran}
\bibliography{references}

\end{document}

%% file: images/autoregressive_pol.tex
% \documentclass[border=0.1cm, 12pt]{standalone}
% \usepackage{tikz}
% \usepackage{amsfonts}
% \usepackage{amsmath,amssymb}
% \usepackage{systeme,mathtools}
% \usetikzlibrary{positioning,arrows.meta,quotes}
% \usetikzlibrary{shapes,snakes}
% \usetikzlibrary{bayesnet}
% \tikzset{>=latex}
% \tikzstyle{plate caption} = [caption, node distance=0, inner sep=0pt,
% below left=5pt and 0pt of #1.south]
% \begin{document}
\begin{tikzpicture}
	\node [circle,draw=black,fill=white,inner sep=0pt,minimum size=0.8cm] (c1) at (-2.6,-2.0) {\scalebox{0.85}[1]{$\boldsymbol{c}_{1}$}};
	\node [circle,draw=black,fill=white,inner sep=0pt,minimum size=0.8cm] (c0) at (-4.2,-2.0) {$\boldsymbol{c}_{0}$};
	\node [circle,draw=black,fill=white,inner sep=0pt,minimum size=0.8cm] (ck) at (0,-2.0) {$\boldsymbol{c}_{k}$};
    \node [circle,draw=black,fill=white,inner sep=0pt,minimum size=0.8cm] (s) at (-4.2,-3.8) {$s$};
    
    \node[text width=0.6cm] (dots) at (-1.26,-2.0) {$\LARGE\cdots$};
    
    %\path (c1) -- node[auto=false]{\ldots} (ck);

    \path [draw,->] (s) edge (c0);
    \path [draw,->] (s) edge (c1);
    \path [draw,->] (s) edge (ck);
    \path [draw,->] (s) edge (dots);
    
    \path [draw,->] (dots) edge (ck);
    \path [draw,->] (c1) edge (dots);
    \path [draw,->] (c0) edge (c1);
    
    % we would add this for autoregressive
    \path [draw,->] (c0) edge[bend left] (dots);
    \path [draw,->] (c0) edge[bend left] (ck);
    \path [draw,->] (c1) edge[bend left] (ck);

\end{tikzpicture}
%\end{document}

%% file: images/mask_dep_pol.tex
% \documentclass[border=0.1cm, 12pt]{standalone}
% \usepackage{tikz}
% \usepackage{amsfonts}
% \usepackage{amsmath,amssymb}
% \usepackage{systeme,mathtools}
% \usetikzlibrary{positioning,arrows.meta,quotes}
% \usetikzlibrary{shapes,snakes}
% \usetikzlibrary{bayesnet}
% \tikzset{>=latex}
% \tikzstyle{plate caption} = [caption, node distance=0, inner sep=0pt,
% below left=5pt and 0pt of #1.south]
% \begin{document}
\begin{tikzpicture}
	\node [circle,draw=black,fill=white,inner sep=0pt,minimum size=0.8cm] (m2) at (-2.6,-0.2) {\scalebox{0.85}[1]{$\boldsymbol{m}_{2}$}};
	\node [circle,draw=black,fill=white,inner sep=0pt,minimum size=0.8cm] (m1) at (-4.2,-0.2) {$\boldsymbol{m}_{1}$};
	\node [circle,draw=black,fill=white,inner sep=0pt,minimum size=0.8cm] (m0) at (-5.8,-0.2) {$\boldsymbol{m}_{0}$};
	%\node [circle,draw=black,fill=white,inner sep=0pt,minimum size=0.8cm] (mk) at (0,-0.2) {$\boldsymbol{m}_{k}$};
	
	\node [circle,draw=black,fill=white,inner sep=0pt,minimum size=0.8cm] (c1) at (-2.6,-2.0) {\scalebox{0.85}[1]{$\boldsymbol{c}_{1}$}};
	\node [circle,draw=black,fill=white,inner sep=0pt,minimum size=0.8cm] (c0) at (-4.2,-2.0) {$\boldsymbol{c}_{0}$};
	\node [circle,draw=black,fill=white,inner sep=0pt,minimum size=0.8cm] (ck) at (0,-2.0) {$\boldsymbol{c}_{k}$};
    
    \node [circle,draw=black,fill=white,inner sep=0pt,minimum size=0.8cm] (s) at (-5.8,-3.8) {$s$};
    
    \node[text width=0.6cm] (dots) at (-1.26,-2.0) {$\LARGE\cdots$};
    \node[text width=0.6cm] (dotsm) at (-1.26,-0.2) {$\LARGE\cdots$};
    
    %\path (c1) -- node[auto=false]{\ldots} (ck);

    \path [draw,->] (s) edge (c0);
    \path [draw,->] (s) edge (c1);
    \path [draw,->] (s) edge (ck);
    \path [draw,->] (s) edge (dots);
    
    %\path [draw,->] (dots) edge (ck);
    \path [draw,->] (c1) edge (m2);
    \path [draw,->] (c0) edge (m1);
    \path [draw,->] (m0) edge (c0);
    \path [draw,->] (m1) edge (c1);
    \path [draw,->] (m2) edge (dots);
    \path [draw,->] (dotsm) edge (ck);
    
    % we would add this for autoregressive
    % \path [draw,->] (c0) edge[bend left] (dots);
    % \path [draw,->] (c0) edge[bend left] (ck);
    % \path [draw,->] (c1) edge[bend left] (ck);

\end{tikzpicture}
% \end{document}